\documentclass{article}
\usepackage{ijcai23}

\usepackage{times}
\usepackage{soul}
\usepackage{url}
\usepackage[hidelinks]{hyperref}
\usepackage[utf8]{inputenc}
\usepackage[small]{caption}
\usepackage{graphicx}
\usepackage{amsmath}
\usepackage{amsthm}
\usepackage{booktabs}
\usepackage{algorithm}
\usepackage{algorithmic}
\usepackage[switch]{lineno}

\usepackage{multirow}
\usepackage{color} 
\usepackage{amssymb}
\usepackage{pifont}
\usepackage{subfigure}
\usepackage{colortbl}
\usepackage{xcolor}
\usepackage{svg}
\usepackage{balance}

\title{Local-Global Transformer Enhanced Unfolding Network for Pan-sharpening}

\author{
Mingsong Li$^1$\and
Yikun Liu$^1$\and
Tao Xiao$^1$\and
Yuwen Huang$^2$\and
and Gongping Yang$^1$\thanks{Corresponding author}
\affiliations
$^1$School of Software, Shandong University, Jinan, China\\
$^2$School of Computer, Heze University, Heze, China
\emails
\{msli, peachxiao\}@mail.sdu.edu.cn,
\{liuyk29, hzxy\_hyw\}@163.com,
gpyang@sdu.edu.cn
}
\begin{document}

\maketitle

\begin{abstract}
\emph{Pan-sharpening} aims to increase the spatial resolution of the low-resolution multispectral (LrMS) image with the guidance of the corresponding panchromatic (PAN) image. Although deep learning (DL)-based pan-sharpening methods have achieved promising performance, most of them have a two-fold deficiency. For one thing, the universally adopted black box principle limits the model interpretability. For another thing, existing DL-based methods fail to efficiently capture local and global dependencies at the same time, inevitably limiting the overall performance. To address these mentioned issues, we first formulate the degradation process of the high-resolution multispectral (HrMS) image as a unified variational optimization problem, and alternately solve its data and prior subproblems by the designed iterative proximal gradient descent (PGD) algorithm. Moreover, we customize a Local-Global Transformer (LGT) to simultaneously model local and global dependencies, and further formulate an LGT-based prior module for image denoising. Besides the prior module, we also design a lightweight data module. Finally, by serially integrating the data and prior modules in each iterative stage, we unfold the iterative algorithm into a stage-wise unfolding network, \textbf{L}ocal-\textbf{G}lobal \textbf{T}ransformer \textbf{E}nhanced \textbf{U}nfolding \textbf{N}etwork (LGTEUN), for the interpretable MS pan-sharpening. Comprehensive experimental results on three satellite data sets demonstrate the effectiveness and efficiency of LGTEUN compared with state-of-the-art (SOTA) methods. The source code is available at \url{https://github.com/lms-07/LGTEUN}.
\end{abstract}

\section{Introduction}

\begin{figure}[!t]
\centering
\includegraphics[width=0.9\linewidth]{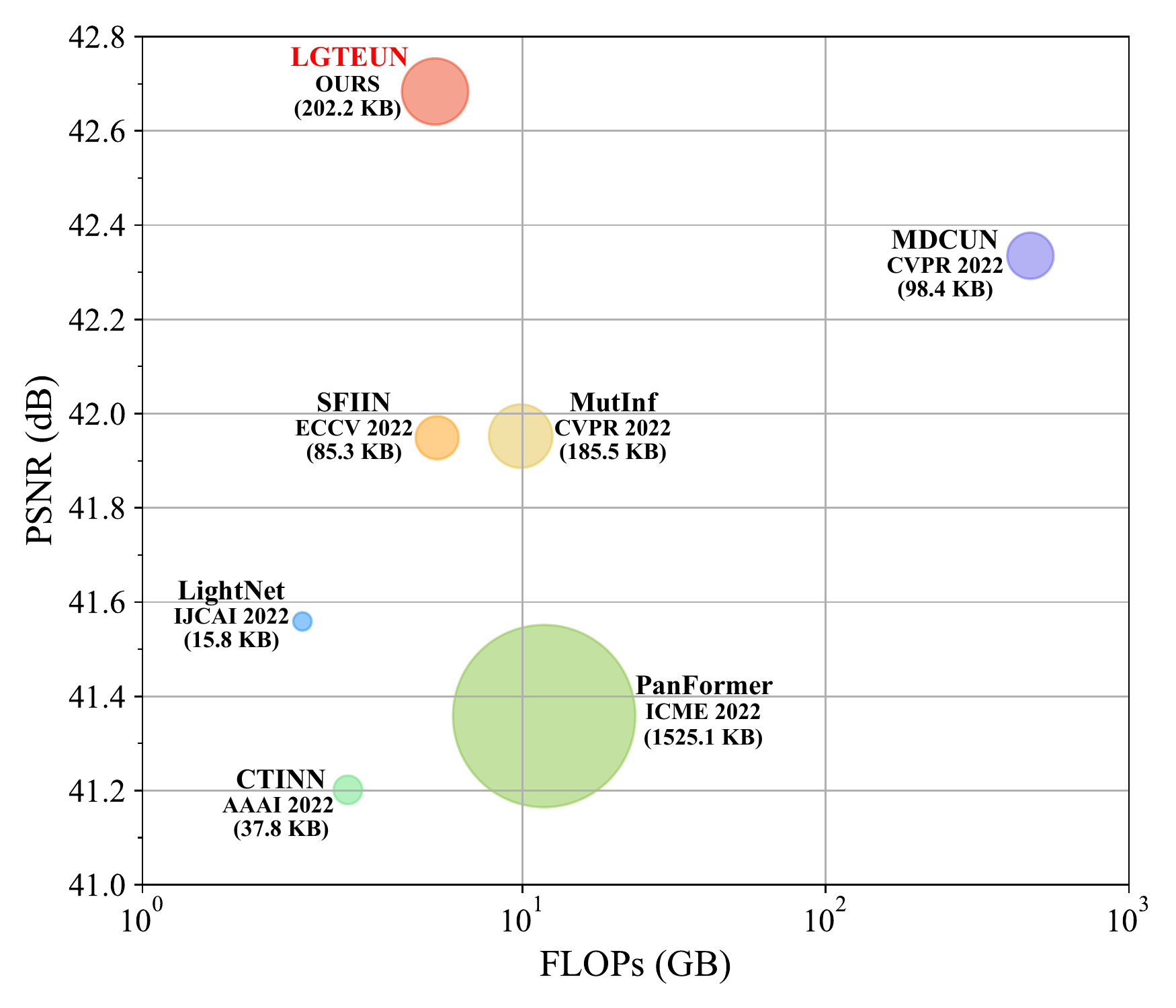}
\vspace{-0.15cm}
\caption{PSNR-Params-FLOPs comparisons between six SOTA DL-based pan-sharpening methods and our LGTEUN on the WorldView-2 satellite data set. The vertical axis is PSNR (model performance), the horizontal axis is FLOPs (computational cost), and the circle radius is Params (model complexity).}
\label{fig1}
\end{figure}

With the development of remote sensing field, multispectral (MS) image is capable of recording more abundant spectral signatures in spectral domain compared with RGB image, and is widely applied in various fields, e.g., environmental monitoring, precision agriculture, and urban development \cite{hardie2004map,fauvel2012advances}. However, due to the inherent trade-off between spatial and spectral resolution, it is hard to directly acquire high-resolution multispectral (HrMS) images. \emph{Pan-sharpening}, a vital yet challenging remote sensing image processing task, aims to produce a HrMS image from the coupled low-resolution multispectral (LrMS) and panchromatic (PAN) images.

Formally, the degradation process of the HrMS image $\mathbf{Z} \in\mathbb{R}^{HW \times B}$ is often expressed as \cite{hardie2004map,xie2019multispectral,dong2021model}:
\begin{equation}\label{eq1}
\begin{split}
     \mathbf{X}=\mathbf{S}\mathbf{Z}+\mathbf{N}_{x},\  \mathbf{Y}=\mathbf{Z}\mathbf{R}+\mathbf{N}_{y},
\end{split}
\end{equation}
where $\mathbf{S} \in\mathbb{R}^{hw \times HW}$ is a linear operator for the spatial blurring and downsampling, $\mathbf{R} \in\mathbb{R}^{B \times 1}$ is the spectral response function of the PAN imaging sensor, and $\mathbf{N}_{x}$ and $\mathbf{N}_{y}$ are the introduced noises during the image acquisition of the LrMS image $\mathbf{X} \in\mathbb{R}^{hw \times B}$ and the PAN image $\mathbf{Y} \in\mathbb{R}^{HW \times 1}$, respectively. Here, $H$ and $h$ ($H>h$), $W$ and $w$ ($W>w$), and $B$ represent the spatial height, the spatial weight, and the number of spectral bands of the corresponding image, respectively. In the past few decades, many methods have been developed in light of the degradation process in Eq. \eqref{eq1}, which can be roughly divided into two categories, i.e., model-based and deep learning (DL)-based.

Typical model-based methods include component substitution (CS) \cite{aiazzi2007improving}, multiresolution analysis (MRA) \cite{liu2000smoothing,king2001wavelet}, and variational optimization (VO) \cite{ballester2006variational,fu2019variational}. These methods rely on prior subjective assumptions in the super-resolving process, and show limited model performance and generalization ability in real scenes. Attracted by the impressive success of DL in various vision tasks \cite{he2016deep,chollet2017xception}, many DL-based methods have been developed for pan-sharpening, especially convolutional neural network (CNN)-based methods. Owing to the outstanding feature representation in hierarchical manner, DL-based methods are capable of directly learning strong priors, and achieve competitive performance, e.g., PanNet \cite{yang2017pannet} and SDPNet \cite{xu2020sdpnet}. 

\textbf{Model Interpretability:} Despite the strong feature extraction ability and encouraging performance improvement, the weak model interpretability is a longstanding deficiency for DL-based methods due to the adopted black box principle. To this end, deep unfolding networks (DUNs) combine merits of both model-based and DL-based methods, and reasonably formulate end-to-end DL models tailored to the investigated pan-sharpening problem employing the theoretical designing philosophy. For instance, Xu \emph{et al.} \cite{xu2021deep} developed the first DUN for pan-sharpening, justifying the generative models for LrMS and PAN images. 

\textbf{Local and Global Dependencies:} Although existing DUNs strengthen the model interpretability towards the investigated pan-sharpening problem, their potential has been far from fully explored. Here, for DUNs, we claim that a competitive denoiser of the image denoising step would sufficiently complement the data step in each iterative stage. However, limited by the local receptive field, most popular CNN-based denoisers pay less attention to global dependencies, which are as important as local dependencies. Furthermore, global transformer, e.g., Vision Transformer (ViT) \cite{dosovitskiy2020image}, can capture global dependencies, obtaining outstanding performance in vision tasks. Yet, global transformer has nontrivial quadratic computational complexity to input image size due to the computation of global self-attention, which inescapably decreases model efficiency. 

Similarly, MDCUN \cite{yang2022memory} employed non-local prior and non-local block \cite{wang2018non} for modeling long-range dependencies, thus showing high computational cost. Besides DUNs, Zhou \emph{et al.} \cite{zhou2022panformer} proposed a modality-specific PanFormer based on Swin Transformer \cite{liu2021swin}. To capture local and long-range dependencies, Zhou \emph{et al.} \cite{zhou2022pan} designed a CNN and transformer dual-branch model, CTINN. However, no matter the serial model, e.g., PanFormer, or the dual-branch model, e.g., CTINN, they both fail to model local and global dependencies in the same layer, which inevitably generates few limitations to the image denoiser or the total pan-sharpening model.

Following the above analysis, in this paper, we develop a transformer-based deep unfolding network, \textbf{L}ocal-\textbf{G}lobal \textbf{T}ransformer \textbf{E}nhanced \textbf{U}nfolding \textbf{N}etwork (LGTEUN), for the interpretable MS pan-sharpening. To be specific, we first formulate a unified variational optimization problem in light of the degenerating observation of pan-sharpening, and design an iterative proximal gradient descent (PGD) algorithm to alternately solve its data and prior subproblems. Second, we elaborate a Local-Global Transformer (LGT) as a prior module for image denoising. The key component in each LGT basic block is its token mixer, the Local-Global Mixer (LG Mixer), which consists of a \emph{local branch} and a \emph{global branch}. The \emph{local branch} calculates local window based self-attention in spatial domain, while the \emph{global branch} extracts global contextual feature representation in frequency domain. Therefore, the LGT-based prior module can simultaneously capture local and global dependencies, and we also design a lightweight data module. Finally, when unfolding the iterative algorithm into the stage-wise unfolding network, LGTEUN, we serially integrate the lightweight data module and the powerful prior module in each iterative stage. Extensive experimental results on three satellite data sets demonstrate the superiority of our method compared with other state-of-the-art (SOTA) methods (as shown in Fig. \ref{fig1}). Our contributions can be summarized as follows:

\textbf{1)} We customize a transformer module LGT as an image denoiser to efficiently model local and global dependencies at the same time and sufficiently mine the potential of the proposed unfolding pan-sharpening framework.

\textbf{2)} We develop an interpretable transformer-based deep unfolding network, LGTEUN. To the best of our knowledge, LGTEUN is the first transformer-based deep unfolding network for the MS pan-sharpening, and LGT is also the first transformer module to perform spatial and frequency dual-domain learning.

\section{Related Work}
In this section, the related deep unfolding networks and transformer-based methods are briefly reviewed.

\subsection{Deep Unfolding Network}
Through integrating merits of both model-based and DL-based methods, deep unfolding networks (DUNs) much improve the interpretability of DL-based models. DUN unfolds the iterative algorithm tailored to the investigated problem, and optimizes the algorithm employing neural modules in an end-to-end trainable manner. DUN has been utilized to solve different low-level vision tasks, including image denoising \cite{mou2022deep}, image compressive sensing \cite{zhang2018ista}, image reconstruction \cite{cai2022degradation}, and image super-resolution \cite{xie2019multispectral,zhang2020deep,dong2021model}. For the discussed MS pan-sharpening, GPPNN \cite{xu2021deep} and MDCUN \cite{yang2022memory} are two representative DUNs. However, restricted by the local receptive field, the adopted CNN-based denoiser in GPPNN pays less attention to global dependencies, which is adverse for reducing copy artifacts. Although MDCUN introduces non-local prior and non-local block to model long-range dependencies, the additional computational cost is heavy. Thus, it is still a crucial issue for DUN to formulate a competitive denoiser of the image denoising step to efficiently capture local and global dependencies and further sufficiently complement the data step in each iterative stage.

\subsection{Transformer}
Originating from language tasks \cite{vaswani2017attention}, transformer has an excellent ability to capture global dependencies, and has been widely applied in various vision tasks, e.g., image classification, object detection, semantic segmentation, and image restoration \cite{dosovitskiy2020image,liu2021swin,liang2021swinir,zhou2022panformer,zhou2022pan}. However, transformer encounters two main issues. \textbf{1)} Global transformer has nontrivial quadratic computation complexity to input image size due to the computation of image-level self-attention. \textbf{2)} Although equipped with considerable-size windows and non-local interactions across windows, local transformer still has difficulties to model image-level global dependencies. Moreover, for pan-sharpening task, transformer-involved models, e.g., \cite{zhou2022panformer} and \cite{zhou2022pan}, fail to process both local and global dependencies at the same time, which inevitably limits the overall performance. 

\section{Method}
\subsection{Model Formulation and Optimization}
Technically, under the maximizing a posterior (MAP) framework, recovering the original HrMS image based on the degradation process in Eq. \eqref{eq1} is a typical ill-posed problem. Generally, the estimation of the HrMS image $\mathbf{Z}$ is implemented by minimizing the following energy function as
\begin{equation}\label{eq2}
\begin{split}
    \bar{\mathbf{Z}}
    &=\underset{\mathbf{Z}}{argmin}\frac{1}{2}{\parallel\mathbf{X}-\mathbf{S}\mathbf{Z}\parallel}^2+\frac{1}{2}{\parallel\mathbf{Y}-\mathbf{Z}\mathbf{R}\parallel}^2+\lambda J(\mathbf{Z}),\\
\end{split}
\end{equation}
where $\frac{1}{2}{\parallel\mathbf{X}-\mathbf{S}\mathbf{Z}\parallel}_2^2$ and $\frac{1}{2}{\parallel\mathbf{Y}-\mathbf{Z}\mathbf{R}\parallel}_2^2$ are the two data fidelity terms coinciding with the degenerating observation, $J(\mathbf{Z})$ is the prior term to constraint the solution space, and $\lambda$ is a trade-off parameter.

Subsequently, proximal gradient descent (PGD) algorithm \cite{beck2009fast} is employed to solve Eq. \eqref{eq2} as an iterative convergence problem, i.e.,
\begin{equation}\label{eq3}
\begin{split}
    \bar{\mathbf{Z}}_k
    &=\underset{\mathbf{Z}}{argmin}\frac{1}{2}{\parallel\mathbf{Z}-(\bar{\mathbf{Z}}_{k-1}-\eta\nabla_f(\bar{\mathbf{Z}}_{k-1})\parallel}^2+\lambda J(\mathbf{Z}),\\
\end{split}
\end{equation}
where $\bar{\mathbf{Z}}_k$ denotes the output of the $k$-th iteration, and $\eta$ is the step size. Here, the data terms oriented differentiable operator $\nabla_f(\bar{\mathbf{Z}}_{k-1})$ is further calculated as
\begin{equation}\label{eq4}
\begin{split}
    \nabla_f(\bar{\mathbf{Z}}_{k-1})
    &=\mathbf{S}^T(\mathbf{S}\bar{\mathbf{Z}}_{k-1}-\mathbf{X})+(\bar{\mathbf{Z}}_{k-1}\mathbf{R}-\mathbf{Y})\mathbf{R}^T.\\
\end{split}
\end{equation}

Moreover, this iterative problem can be addressed by alternately solving its data subproblem at the gradient descent step (Eq. \eqref{eq5}) and its prior subproblem at the proximal mapping step (Eq. \eqref{eq6}). In detail,
\begin{equation}\label{eq5}
\begin{split}
    \bar{\mathbf{Z}}_{k-\frac{1}{2}}
    &=\bar{\mathbf{Z}}_{k-1}-\eta\nabla(\bar{\mathbf{Z}}_{k-1}),\\
\end{split}
\end{equation}
\begin{equation}\label{eq6}
\begin{split}
    \bar{\mathbf{Z}}_{k}
    &=prox_{\eta,J}(\bar{\mathbf{Z}}_{k-\frac{1}{2}}),\\
\end{split}
\end{equation}
where $prox_{\eta,J}$ represents the proximal operator dependent on the prior term $J(\cdot)$. In this way, the PGD algorithm utilizes a few iterations to alternately update $\bar{\mathbf{Z}}_{k-\frac{1}{2}}$ and $\bar{\mathbf{Z}}_{k}$ until convergence. In particular, from a Bayesian perspective, the solution of the prior subproblem Eq. \eqref{eq6} corresponds to a Gaussian denoising problem with noise level $\sqrt{\lambda}$ \cite{chan2016plug,zhang2020deep,mou2022deep,cai2022degradation}. In this work, we elaborate a transformer-based denoiser to approximate the proximal operator $prox_{\eta,J}$, which prominently facilitates the denoising capability and further sufficiently complements the data step in each iterative stage.

\subsection{Deep Unfolding Network}
Through unfolding the iterative PGD algorithm, as illustrated in Fig. \ref{fig2}, we develop our \textbf{L}ocal-\textbf{G}lobal \textbf{T}ransformer \textbf{E}nhanced \textbf{U}nfolding \textbf{N}etwork (LGTEUN). The LGTEUN is comprised of several stages. Each stage contains a lightweight CNN-based data module $\mathcal{D}$ and a powerful transformer-based prior module $\mathcal{P}$, corresponding to the data subproblem at the gradient descent step (Eq. \eqref{eq5}) and the prior subproblem at the proximal mapping step (Eq. \eqref{eq6}) in each iteration, respectively.

\subsubsection{Data Module $\mathcal{D}$} 
To approximate the closed-form solution of the data subproblem at the gradient descent step (Eq. \eqref{eq5}), we design a lightweight CNN-based data module, i.e.,
\begin{equation}\label{eq7}
\begin{split}
    \bar{\mathbf{Z}}_{k-\frac{1}{2}}=\mathcal{D}(\bar{\mathbf{Z}}_{k-1}, \mathbf{X}, \mathbf{Y}, \eta_{k-1}).\\
\end{split}
\end{equation}
Specifically, as shown in Fig. \ref{fig2} (a), the data module of the $k$-th stage takes the out of the $k-1$-th stage $\bar{\mathbf{Z}}_{k-1}$, the LrMS image\footnote{As an explanatory instance, the dimension of $\mathbf{X}$ is $HW\times B$ in mathematical formalization like Eq. \eqref{eq1}, and $H\times W\times B$ in programming implementation here.} $\mathbf{X}$, the PAN image $\mathbf{Y}$, and the stage-specific learnable step size $\eta_{k-1}$ as its module inputs. What's more, the matrix $\mathbf{S}$ is implemented by two downsampling units, and each unit consists of a downsampling operation and a 3$\times$3 depth convolution (Conv) layer \cite{chollet2017xception}. Similarly, the transposed matrix $\mathbf{S}^T$ is performed by two upsampling units. Besides, one point Conv \cite{chollet2017xception} is utilized as the matrix $\mathbf{R}$ to reduce channels from $B$ to $1$, and another point Conv is utilized as the matrix $\mathbf{R}^T$ for the corresponding inverse channel increase.

\subsubsection{Prior Module $\mathcal{P}$}
Considering the designing of the denoiser at the image denoising step, previous DUNs are mainly based on CNN, presenting limitations in capturing global dependencies. Here, as the first transformer-based image denoiser in the MS pan-sharpening oriented DUN, we dedicate significant efforts to craft a Local-Global Transformer (LGT) as the key denoising prior module $\mathcal{P}$, i.e.,
\begin{equation}\label{eq8}
\begin{split}
    \bar{\mathbf{Z}}_k=\mathcal{P}(\bar{\mathbf{Z}}_{k-\frac{1}{2}}).\\
\end{split}
\end{equation}

\begin{figure*}[!t]
\centering
\includegraphics[width=0.95\linewidth]{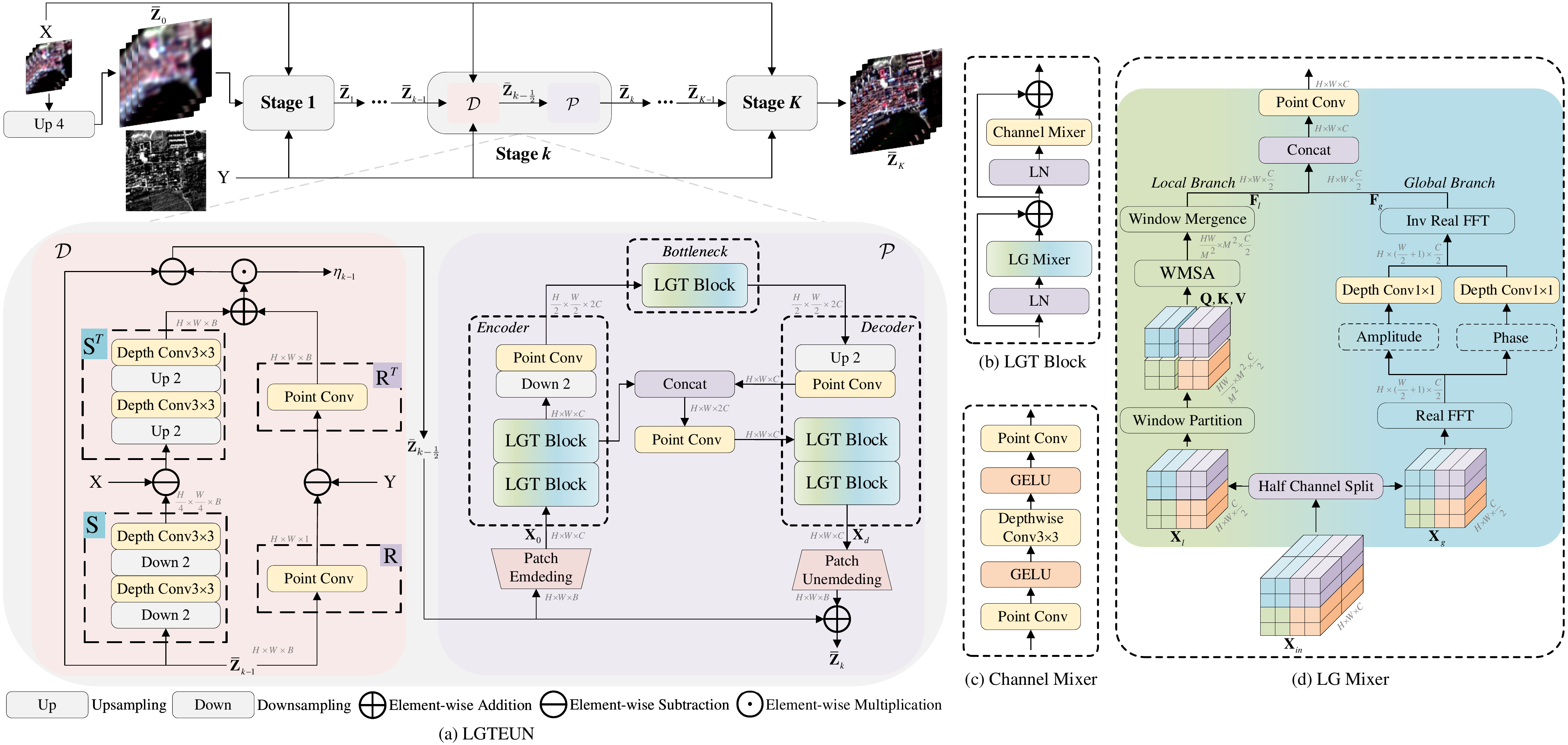}
\vspace{-0.15cm}
\caption{Illustration of the proposed LGTEUN. (a) The overall architecture of LGTEUN with $K$ stages and details of the $k$-th stage. The lightweight CNN-based data module $\mathcal{D}$ and the powerful transformer-based prior module $\mathcal{P}$ in each stage correspond to the data and prior subproblems in an iteration of the PGD algorithm. (b) Components of an LGT block. (c) The adopted channel mixer. (d) The key LG Mixer is comprised of a \emph{local branch} and a \emph{global branch}.}
\label{fig2}
\end{figure*}

\paragraph{Overall Architecture of LGT.}
As depicted in Eq. \eqref{eq8} and Fig. \ref{fig2} (a), given the output $\bar{\mathbf{Z}}_{k-\frac{1}{2}}\in\mathbb{R}^{H\times W \times B}$ of $\mathcal{D}$ as the module input, LGT applies a U-shaped structure mainly constituted by a series of basic LGT Blocks, and outputs $\bar{\mathbf{Z}}_k\in\mathbb{R}^{H\times W \times B}$ as the module output. Concretely, LGT first uses a patch embedding layer to split the intermediate image $\bar{\mathbf{Z}}_{k-\frac{1}{2}}$ into non-overlapping patch tokens and further produces the embedded feature $\mathbf{X}_0\in\mathbb{R}^{H\times W \times C}$. Second, an encoder-bottleneck-decoder structure extracts the  discriminative feature representation $\mathbf{X}_d\in\mathbb{R}^{H\times W \times C}$ from $\mathbf{X}_0$. In particular, the encoder and decoder both contain two LGT blocks and a resizing unit, and the bottleneck has a single LGT block. In each resizing unit, the downsampling or upsampling operation is responsible for resizing spatial resolution, and a point Conv changes the channel dimension accordingly. Finally, a patch unembedding layer is employed to project $\mathbf{X}_d$ to $\bar{\mathbf{Z}}_k$. Here, note that the patch size is set as 1, thus the original pixel vectors in $\bar{\mathbf{Z}}_{k-\frac{1}{2}}$ act as the discussed patch tokens for finer local and global token mixing.

In Fig. \ref{fig2} (b), each LGT block consists of a layer normalization (LN), a Local-Global Mixer (LG Mixer) for mixing the spatial information, a LN, and a channel mixer in order. As illustrated in Fig. \ref{fig2} (c), the channel mixer is a depthwise Conv \cite{chollet2017xception} based neural module for efficient channel mixing. Specifically, the LG Mixer as the token mixer is the key component in each LGT block, and Fig. \ref{fig2} (d) depicts the LG Mixer of the first LG Block in the encoder. For convenience, let $\mathbf{X}_{in}\in\mathbb{R}^{H\times W \times C}$ represent the input feature map of our LG Mixer, which is further split into two equal parts $\mathbf{X}_{l}\in\mathbb{R}^{H\times W \times \frac{C}{2}}$ and $\mathbf{X}_{g}\in\mathbb{R}^{H\times W \times \frac{C}{2}}$ along the channel dimension. Then $\mathbf{X}_{l}$ and $\mathbf{X}_{g}$ are assigned to a \textit{local branch} and a \textit{global branch}, respectively. The \emph{local branch} models local dependencies by computing local window based self-attention in spatial domain, while the \emph{global branch} captures global dependencies by mining global contextual feature representation in frequency domain. By concatenating the output of the \emph{local branch} $\mathbf{F}_l \in\mathbb{R}^{H\times W \times \frac{C}{2}}$ and that of the \emph{global branch} $\mathbf{F}_g \in\mathbb{R}^{H\times W \times \frac{C}{2}}$, the local and global dependencies are simultaneously captured in our LG Mixer.

\paragraph{Local Branch.} 
The \emph{local branch} calculates local window based multi-head self-attention (WMSA) in spatial domain. In detail, as shown in the left path of Fig. \ref{fig2} (d), $\mathbf{X}_{l}$ is first partitioned into non-overlapping windows, and each window contains $M \times M$ patch tokens. Then the window-specific feature map with $\frac{HW}{M^2} \times M^2 \times \frac{C}{2}$ dimension is obtained by simply reshaping. Subsequently, three feature embeddings $\mathbf{Q}$, $\mathbf{K}$, and $\mathbf{V}\in\mathbb{R}^{\frac{HW}{M^2} \times M^2 \times \frac{C}{2}}$ are generated through a point Conv based linear projection. Furthermore, $\mathbf{Q}$, $\mathbf{K}$, and $\mathbf{V}$ are channel-wise divided into $h$ heads, i.e., $\mathbf{Q}=[\mathbf{Q}^1,...,\mathbf{Q}^h]$, $\mathbf{K}=[\mathbf{K}^1,...,\mathbf{K}^h]$, and $\mathbf{V}=[\mathbf{V}^1,...,\mathbf{V}^h]$. Each head contains $d=\frac{C}{2h}$ channels, and Fig. \ref{fig2} (d) only presents the circumstance with $h=1$ for simplification. More importantly, the WMSA map is computed as 

\begin{equation}\label{eq9}
\begin{split}
    \mathbf{F}^i_a=Softmax(\frac{\mathbf{Q}^i{\mathbf{K}^i}^T}{\sqrt{d}}+\mathbf{P}^i)\mathbf{V}^i,\ \ i=1,...,h,\\
\end{split}
\end{equation}
where $\mathbf{P}^i\in\mathbb{R}^{M^2 \times M^2}$ is the learnable position embedding. At last, for the feature map $\mathbf{F}_a$, we channel-wise concatenate its $h$ heads and spatially merge its $\frac{HW}{M^2}$ windows to yield the branch output $\mathbf{F}_l$.

\paragraph{Global Branch.}
The \emph{global branch} extracts global contextual feature representation in frequency domain based on the nature of Fourier transformation. To be specific, according to spectral convolution theorem in Fourier theory \cite{frigo1998fftw,chi2020fast,zhao2022fractional,zhou2022spatial,zhou2022deep}, feature learning in frequency spectral domain has the image-wide receptive field by channel-wise Fourier transformation. Besides, point-wise multiplications in frequency domain correspond to convolutions in spatial domain. These properties provide vital theoretical guidances of our \emph{global branch}.

Formally, 2D discrete Fourier transform (DFT) first converts $\mathbf{X}_{g}$ from spatial domain to Fourier frequency domain as the complex component $\mathcal{F}(\mathbf{X}_{g})$, i.e.,
\begin{equation}\label{eq10}
\begin{split}
    \mathcal{F}(\mathbf{X}_{g})(u,v)
    &=\frac{1}{\sqrt{HW}}\sum\limits_{h=0}^{H-1}\sum\limits_{w=0}^{W-1}\mathbf{X}_{g}(h,w)e^{-j2\pi(\frac{h}{H}u+\frac{w}{W}v)},
\end{split}
\end{equation}
where $u$ and $v$ are frequency components. Here, $\mathcal{F}(\mathbf{X}_{g}) \in\mathbb{C}^{H \times (\frac{W}{2}+1) \times \frac{C}{2}}$ is produced in light of the conjugate symmetry property of 2D DFT for our real input $\mathbf{X}_{g}$, and $\mathbb{C}$ denotes complex domain. Besides, the inverse 2D DFT is accordingly represented as $\mathcal{F}^{-1}(\cdot)$. Then based on the real part $R(\mathbf{X}_{g})$ and the imaginary part $I(\mathbf{X}_{g})$ of $\mathcal{F}(\mathbf{X}_{g})$, the amplitude component $\mathcal{A}(\mathbf{X}_{g})$ and the phase component $\mathcal{P}(\mathbf{X}_{g})$ are further expressed as
\begin{equation}\label{eq11}
\begin{split}
    \mathcal{A}(\mathbf{X}_{g})(u,v)
    &=\sqrt{R^2(\mathbf{X}_{g})(u,v)+I^2(\mathbf{X}_{g})(u,v)},
\end{split}
\end{equation}
\begin{equation}\label{eq12}
\begin{split}
    \mathcal{P}(\mathbf{X}_{g})(u,v)
    &=arctan[\frac{I(\mathbf{X}_{g})(u,v)}{R(\mathbf{X}_{g})(u,v)}].
\end{split}
\end{equation}

Furthermore, two independent 1$\times$1 depth Convs are utilized for feature learning in frequency domain, and the inverse 2D DFT $\mathcal{F}^{-1}(\cdot)$ is applied to recompose the feature representations of the amplitude and phase components back to spatial domain. In detail,
\begin{equation}\label{eq13}
\begin{split}
    \mathbf{F}_g
    &=\mathcal{F}^{-1}(DConv(\mathcal{A}(\mathbf{X}_{g})), DConv(\mathcal{P}(\mathbf{X}_{g}))),
\end{split}
\end{equation}
where $DConv$ represents the applied 1$\times$1 depth Conv. In fact, to improve module efficiency, the 2D DFT and the inverse 2D DFT are computed by the 2D real fast Fourier transform (rFFT) and the inverse 2D rFFT, which can be implemented by $\textit{torch.rfft2}$ and $\textit{torch.irfft2}$ in PyTorch programming framework, respectively. The flowchart of our \emph{global branch} is depicted in the right path of Fig. \ref{fig2} (d). 

\section{Experiments}
\subsection{Data Sets and Evaluation Metrics}
For the MS pan-sharpening, an 8-band MS data set acquired by the WorldView-3 sensor $\footnote{\url{https://www.l3harris.com/all-capabilities/high-resolution-satellite-imagery}\label{fn1}}$ and two 4-band MS data sets acquired by WorldView-2 $\textsuperscript{\ref{fn1}}$ and GaoFen-2 sensors are adopted for experimental analysis. Due to the unavailability of ground-truth (GT) images for training, following Wald's protocol \cite{wald1997fusion}, we employ downsampling operations to produce a reduced-resolution data set for each satellite sensor. Each data set is further split into non-overlapping subsets for training (about 1000 LrMS/PAN/GT image pairs) and testing (about 140 LrMS/PAN/GT image pairs). The spatial sizes of LrMS, PAN, and GT images are $32\times32$, $128\times128$, and $128\times128$, respectively. In addition, we only adopt upsampling operations to produce a full-resolution data set with 120 LrMS/PAN/GT image pairs for the WorldView-3 satellite sensor.

For image quality assessment (IQA), five popular metrics are applied for the reduced-resolution test, i.e., PSNR, SSIM, Q-index, SAM, and ERGAS, and three common non-reference metrics are employed for the full-resolution test, i.e., $D_\lambda$, $D_S$, and QNR. Besides, the inference time, parameters (Params), and floating-point operations (FLOPs) are utilized for model efficiency analysis.

% stage test
\begin{table}[!t]
\caption{Performance and efficiency of LGTEUN with different numbers of stages $K$ on WorldView-3 and WorldView-2 satellite data sets.}
\centering
\scriptsize
\renewcommand{\arraystretch}{1.2}
\begin{tabular}{c|c| cccc}
\toprule
\multicolumn{1}{c}{Data Set} & Metric &Stage 1 &Stage 2 &Stage 3 &Stage 4  \\
\hline
\hline
\multirow{8}{*}{WorldView-3}
&PSNR$\uparrow$   &32.0339 &\textcolor{red}{32.2188} &\textcolor{blue}{32.068} &32.0042  \\
&SSIM$\uparrow$   &0.9532  &\textcolor{red}{0.9545}  &\textcolor{blue}{0.9535} &0.9527  \\
&Q8$\uparrow$  &0.9481  &\textcolor{red}{0.9494}  &\textcolor{blue}{0.9487} &0.9480  \\
&SAM$\downarrow$  &\textcolor{blue}{0.0605}  &\textcolor{blue}{0.0605}  &\textcolor{red}{0.0603} &0.0612\\
&ERGAS$\downarrow$&2.6765  &\textcolor{red}{2.6286}  &\textcolor{blue}{2.6678} &2.6898\\
\cline{2-6}
&Time (s/img)     &\textcolor{red}{0.0070} &\textcolor{blue}{0.0133} &0.0205 &0.0262\\
&Params (KB)      &\textcolor{red}{270.2}  &\textcolor{blue}{540.0}  &809.9  &1079.7\\
&FLOPs (GB)       &\textcolor{red}{9.52}   &\textcolor{blue}{19.04}  &28.56  &38.08\\
\hline
\multirow{8}{*}{WorldView-2}
&PSNR$\uparrow$   &\textcolor{blue}{42.600} &\textcolor{red}{42.6837} &42.4771 &42.1634  \\
&SSIM$\uparrow$   &\textcolor{blue}{0.9784} &\textcolor{red}{0.9786}  &0.9781  &0.9767  \\
&Q4$\uparrow$     &\textcolor{blue}{0.8398} &\textcolor{red}{0.8415}  &0.8383  &0.8329  \\
&SAM$\downarrow$  &\textcolor{blue}{0.0209} &\textcolor{red}{0.0208}  &0.0213  &0.0222\\
&ERGAS$\downarrow$&\textcolor{blue}{0.9358} &\textcolor{red}{0.928}   &0.9573  &0.9787\\
\cline{2-6}
&Time (s/img)     &\textcolor{red}{0.0065} &\textcolor{blue}{0.0137}  &0.0204  &0.0254\\
&Params (KB)      &\textcolor{red}{101.2}  &\textcolor{blue}{202.2}   &303.2   &404.2\\
&FLOPs (GB)       &\textcolor{red}{2.57}   &\textcolor{blue}{5.14}    &7.71    &10.28\\
\bottomrule
\end{tabular}
\label{tab1}
\end{table}

% reduced test
\begin{table*}[!t]
\caption{Quantitative comparison of different methods on WorldView-3, WorldView-2, and GaoFen-2 satellite data sets.}
\centering
\tiny
\renewcommand{\arraystretch}{1.2}
\begin{tabular}{c ccccc ccccc ccccc}
\toprule
\multirow{2.5}{*}{Method}&\multicolumn{5}{c}{WorldView-3} &\multicolumn{5}{c}{WorldView-2} &\multicolumn{5}{c}{GaoFen-2}\\
\cmidrule(lr){2-6}  \cmidrule(lr){7-11} \cmidrule(lr){12-16}
  &PSNR$\uparrow$ &SSIM$\uparrow$ &Q8$\uparrow$ &SAM$\downarrow$ &ERGAS$\downarrow$   &PSNR$\uparrow$ &SSIM$\uparrow$ &Q4$\uparrow$ &SAM$\downarrow$ &ERGAS$\downarrow$         &PSNR$\uparrow$ &SSIM$\uparrow$ &Q4$\uparrow$ &SAM$\downarrow$ &ERGAS$\downarrow$ \\
\hline
\hline
GSA     &22.5164 &0.6343 &0.5742 &0.1106 &7.8267   &33.5975 &0.8899 &0.5681 &0.0573 &2.5402    &36.0557 &0.8838 &0.5517 &0.0641 &3.5758\\
SFIM    &21.4154 &0.5415 &0.4525 &0.1147 &8.8553   &32.6334 &0.8728 &0.5159 &0.0597 &3.1919    &34.7715 &0.8572 &0.4584 &0.0657 &4.2073 \\
Wavelet &21.4464 &0.5656 &0.5271 &0.1503 &9.1545   &32.1992 &0.8500 &0.4577 &0.0638 &3.3799    &33.9208 &0.8197 &0.4033 &0.0695 &4.6445\\
\hline
PanFormer &30.4772 &0.9368 &0.9316 &0.0672 &3.1830  &41.3581 &0.9731 &0.8236 &0.0241 &1.0617    &44.8540 &0.9805 &0.8865 &0.0271 &1.3334\\
CTINN     &31.8564 &0.9518 &0.9460 &0.0660 &2.7421   &41.2015 &0.9735 &0.8149 &0.0246 &1.0880   &44.2942 &0.9784 &0.8716 &0.0293 &1.4148  \\
LightNet  &\textcolor{blue}{32.0018} &\textcolor{blue}{0.9525} &\textcolor{blue}{0.9472} &0.0639 &\textcolor{blue}{2.6853}   &41.5589 &0.9739 &0.8220 &0.0237 &1.0382   &44.6876 &0.9787 &0.8741 &0.0279 &1.3510\\
SFIIN     &31.6587 &0.9492 &0.9435 &0.0652 &2.8016   &41.9489 &0.9752 &0.8108 &0.0229 &1.0084   &44.7248 &0.9802 &0.8721 &0.0280 &1.3361\\
MutInf    &31.8298 &0.9523 &0.9469 &\textcolor{blue}{0.0636} &2.7526   &41.9522 &0.9760 &0.8258 &0.0227 &1.0153   &44.8305 &0.9800 &0.8836 &0.0277 &1.3394\\
MDCUN     &31.2978 &0.9429 &0.9363 &0.0661 &2.9295   &\textcolor{blue}{42.3351} &\textcolor{blue}{0.9772} &\textcolor{blue}{0.8370} &\textcolor{blue}{0.0216} &\textcolor{blue}{0.9638}   &\textcolor{blue}{45.5677} &\textcolor{blue}{0.9825} &\textcolor{blue}{0.8915} &\textcolor{blue}{0.0252} &\textcolor{blue}{1.2249}\\ 
\hline
LGTEUN   &\textcolor{red}{32.2188} &\textcolor{red}{0.9545} &\textcolor{red}{0.9494} &\textcolor{red}{0.0605} &\textcolor{red}{2.6286}   &\textcolor{red}{42.6837} &\textcolor{red}{0.9786} &\textcolor{red}{0.8415} &\textcolor{red}{0.0208} &\textcolor{red}{0.9280}  &\textcolor{red}{45.8364} &\textcolor{red}{0.9840} &\textcolor{red}{0.8973} &\textcolor{red}{0.0247} &\textcolor{red}{1.1824}\\
\bottomrule
\end{tabular}
\label{tab2}
\end{table*}

\begin{figure*}[!t]
\centering
\includegraphics[width=0.925\linewidth]{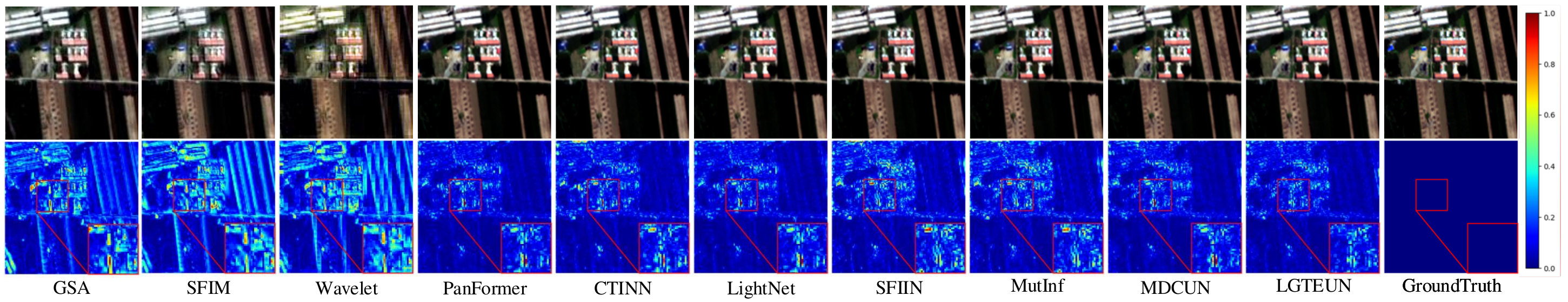}
\vspace{-0.25cm}
\caption{Qualitative comparison of different methods on the WorldView-2 satellite data set.}
% \vspace{-0.15cm}
\label{fig3}
\end{figure*}

\subsection{Implementation Details}
\paragraph{Training Setting.}
The end-to-end training of LGTEUN is supervised by mean absolute error (MAE) loss between the network output $\bar{\mathbf{Z}}_{K}$ and the GT HrMS image. It trains $130$ epochs for the 8-band data set, and $1000$ epochs for the two 4-band data sets. The Adam optimizer with $\beta_1=0.9$ and $\beta_2=0.999$ is employed for model optimization, and the batch size is set as $4$. The initial learning rate is $1.5\times10^{-3}$, and decays by $0.85$ every $100$ epochs. All the experiments are conducted in PyTorch framework with a single NVIDIA GeForce GTX 3090 GPU. For clear comparisons, \textcolor{red}{red} color highlights the best results while \textcolor{blue}{blue} color the second-best in the following suitable table results.
\paragraph{Structure Setting.}
As shown in Fig. \ref{fig2} (a), $\bar{\mathbf{Z}}_{0}$ is initialized by directly upsampling the LrMS image $\mathbf{X}$ with a scaling factor $4$. In LGTEUN, the data module $\mathcal{D}$ shares parameters across stages, while the prior module maintains independence. The channel number $C$ is set as $4B$ for all the data sets. Additionally, all the downsampling or upsampling operation is implemented by the bicubic interpolation. 

Besides, in the $\emph{local branch}$ of each LGT block, the size of each local window $M$ is $8$, and the number of heads is $2$. Moreover, in Tab. \ref{tab1}, we explore the impact of the number of iterative stages $K=1,2,3,4$ from performance and efficiency viewpoints on 8-band WorldView-3 and 4-band WorldView-2 satellite scenes. According to the results in Tab. \ref{tab1}, it is clear that LGTEUN reaches the optimal performance with $2$ stages, and the model efficiency gradually decreases as $K$ increases. Hence, the number of stages is chosen as $2$ for better performance and efficiency balance.

\subsection{Comparison with SOTA Methods}
To comprehensively evaluate the effectiveness and efficiency of the proposed method for the MS pan-sharpening, we compare our LGTEUN with three model-based methods, i.e., GSA \cite{aiazzi2007improving}, SFIM \cite{liu2000smoothing}, and Wavelet \cite{king2001wavelet}, and six SOTA DL-based methods, i.e., PamFormer \cite{zhou2022panformer}, CTINN \cite{zhou2022panformer}, LightNet \cite{chen2022spanconv}, SFIIN \cite{zhou2022spatial}, MutInf \cite{zhou2022mutual}, and MDCUN \cite{yang2022memory}. Besides, all the compared methods are implemented in light of the corresponding paper and source code. It is noteworthy that all the six DL-based methods are the most recent algorithms for the MS pan-sharpening.

\paragraph{Quantitative Comparison.}
Tab. \ref{tab2} reports the comparison results of all the discussed ten methods on all the three satellite data sets. Specifically, on all the three data sets, the three model-based methods show limited model performances and generalization abilities, and the DL-based methods obtain more competitive results. More importantly, among all the considered methods on all the three data sets, our proposed LGTEUN always achieves the best results in all the five IQA metrics with distinct performance improvements. For instance, our LGTEUN outperforms the second-best method by 0.2170 dB, 0.3486 dB, and 0.2687 dB in PSNR on WorldView-3, WorldView-2, and GaoFen-2 data sets, respectively, which indicates the superiority of our proposed method.

% full-resolution test
\begin{table}[!t]
\caption{Full-resolution test of different methods on the WorldView-3 satellite data set.}
\centering
\scriptsize
\renewcommand{\arraystretch}{1.2}
\begin{tabular}{c ccc}
\toprule
\multirow{2.5}{*}{Method}&\multicolumn{3}{c}{Full-resolution Test} \\
\cmidrule(lr){2-4} 
 &$D_{\lambda}$$\downarrow$ &$D_{S}$$\downarrow$ &QNR$\uparrow$      \\
\hline
\hline
GSA        &\textcolor{red}{0.0094} &0.1076 &0.8839    \\
SFIM       &\textcolor{red}{0.0094} &0.1061 &0.8854     \\
Wavelet    &0.0552 &0.1330 &0.8193    \\
\hline
PanFormer   &0.0191 &0.0416 &0.9400     \\
CTINN       &\textcolor{blue}{0.0123} &0.0442 &0.9440        \\
LightNet    &0.0185 &\textcolor{red}{0.0282} &\textcolor{red}{0.9539}    \\
SFIIN       &0.0198 &0.0352 &0.9457    \\
MutInf      &0.0163 &0.0420 &0.9423   \\
MDCUN       &0.0747 &0.1673 &0.7708    \\
\hline
LGTEUN      &0.0162 &\textcolor{blue}{0.0310} &\textcolor{blue}{0.9532}   \\
\bottomrule
\end{tabular}
\label{tab3}
\end{table}

% efficiency comparison
\begin{table*}[!t]
\centering
\scriptsize
\caption{Efficiency comparison of different methods on WorldView-3 and GaoFen-2 satellite data sets.}
\renewcommand{\arraystretch}{1.15}
\begin{tabular}{c|c| ccc cccccc c}
\toprule
\multicolumn{1}{c}{Data Set} &Metric  &GSA &SFIM &Wavelet & PanFormer &CTINN &LightNet  &SFIIN &MutInf &MDCUN & LGTEUN      \\
\hline
\hline
\multirow{3}{*}{WorldView-3} 
& Time (s/img)     &0.0482 &0.0591 &0.0562     &0.0160       &0.0426     &\textcolor{red}{0.0019}      &0.0529       &0.1083       &0.1747      &\textcolor{blue}{0.0133}           \\
& Params (KB)          & --   &-- &--  & 1532.8   & \textcolor{blue}{38.3}    & \textcolor{red}{16.3}    & 85.8     & 185.8    & 140.9     & 540.0   \\
& FLOPs (GB)           & --   &-- &--  & 11.92   & \textcolor{blue}{2.68}  & \textcolor{red}{2.02}   & 5.25   & 9.87   & 479.54     & 19.04    \\
\hline
\multirow{3}{*}{GaoFen-2} 
& Time (s/img)     &0.0216 &0.0301 &0.0271     &0.0257       &0.0431     &\textcolor{red}{0.0017}      &0.0528       &0.1141       &0.1017      &\textcolor{blue}{0.0129}           \\
& Params (KB)          & --   &-- &--  & 1530.3   & \textcolor{blue}{37.8}     & \textcolor{red}{15.8}    & 85.3     & 185.5    & 98.3     & 202.2   \\
& FLOPs (GB)           & --   &-- &--  & 11.77   & \textcolor{blue}{2.65}  & \textcolor{red}{1.95}   & 5.22    & 9.85   & 473.19     & 5.14    \\
\bottomrule
\end{tabular}
\vspace{-0.15cm}
\label{tab4}
\end{table*}

% ablation study
\begin{table*}[!t]
\caption{Ablation study on the WorldView-3 satellite data set.}
\centering
\scriptsize
\renewcommand{\arraystretch}{1.2}
\begin{tabular}{cc ccccc ccc}
\toprule
\multicolumn{2}{c}{Setting}  &\multicolumn{5}{c}{Reduced-resolution Test} &\multicolumn{3}{c}{Full-resolution Test} \\
\cmidrule(lr){1-2} \cmidrule(lr){3-7}  \cmidrule(lr){8-10}
\emph{Local Branch} & \emph{Global Branch} &PSNR$\uparrow$ &SSIM$\uparrow$ &Q8$\uparrow$ &SAM$\downarrow$ &ERGAS$\downarrow$   &$D_{\lambda}$$\downarrow$ &$D_{S}$$\downarrow$ &QNR$\uparrow$ \\
\hline
\hline
\ding{55}&\checkmark  &31.9309 &0.9519 &0.9468 &0.0636 &2.7102  &0.0177 &0.0364 &0.9465 \\
\checkmark&\ding{55}  &31.9742 &0.9525 &0.9468 &0.0618 &2.7029   &0.0170 &0.0349 &0.9486 \\
\cellcolor{lightgray}\checkmark&\cellcolor{lightgray}\checkmark &\cellcolor{lightgray}\textcolor{red}{32.2188} &\cellcolor{lightgray}\textcolor{red}{0.9545} &\cellcolor{lightgray}\textcolor{red}{0.9494} &\cellcolor{lightgray}\textcolor{red}{0.0605} &\cellcolor{lightgray}\textcolor{red}{2.6286} &\cellcolor{lightgray}\textcolor{red}{0.0162} &\cellcolor{lightgray}\textcolor{red}{0.0310} &\cellcolor{lightgray}\textcolor{red}{0.9532} \\
\bottomrule
\end{tabular}
\label{tab5}
\end{table*}

\paragraph{Qualitative Comparison.}
Fig. \ref{fig3} illustrates the qualitative results of a typical sample from the WorldView-2 data set, including the paired output pan-sharpening image and the corresponding MAE residual image of each discussed method. In Fig. \ref{fig3}, compared with the other nine methods, the proposed LGTEUN exhibits a more visually pleasing result with minor spectral and spatial distortions. In particular, the residual image of our method has fewer artifacts than any other method, especially in the zoom-in region. Here, we can reasonably infer that the advanced performance of the LGTEUN benefits from the designed PGD algorithm based stage iterations and the excellent capability of simultaneously capturing local and global dependencies.

\paragraph{Full-resolution Test.}
To further measure the model performance in the full-resolution scene, we conduct a full-resolution test on the full-resolution WorldView-3 data set. As reported in Tab. \ref{tab3}, the proposed method also obtains competitive results, i.e., second-best results in $D_S$ and QNR and the third-best result in $D_\lambda$. On the contrary, MDCUN \cite{yang2022memory} exhibits slightly limited results. 

\paragraph{Efficiency Comparison.}
As for efficiency comparison, Tab. \ref{tab4} presents exhaustive investigations about inference efficiency (the inference time), model complexity (Params), and computational cost (FLOPs) of all the ten methods on WorldView-3 and GaoFen-2 data sets, and Fig. \ref{fig1} illustrates the unified PSNR-Params-FLOPs comparisons of all the DL-based methods on the WorldView-2 scenario. Specifically, from Tab. \ref{tab4}, the proposed LGTEUN has excellent inference efficiency and promising computational cost. Besides, further considering the outstanding model performance in Fig. \ref{fig1}, our LGTEUN achieves an impressive performance-efficiency balance.

\subsection{Analysis and Discussion}

\paragraph{Ablation Study.}
In this subsection, we perform an ablation study towards our elaborated LGT in the prior module $\mathcal{P}$. Specifically, on the WorldView-3 data set, two break-down ablation tests are conducted to explore and validate the corresponding contributions of its key $\emph{local}$ and $\emph{global branches}$.

\textbf{Local Branch:} For one thing, the $\emph{local branch}$ models local dependencies by computing local window based self-attention in spatial domain. As reported in Tab. \ref{tab5}, the $\emph{local branch}$ brings obvious performance gains for both reduced-resolution and full-resolution tests. For example, our LGTEUN improves 0.2879 dB in PSNR and 0.0816 in ERGAS for the reduced-resolution test, and 0.0067 in QNR for the full-resolution test, respectively.

\textbf{Global Branch:} For another thing, the \emph{global branch} captures global dependencies by mining global contextual feature representation in frequency domain. From Tab. \ref{tab5}, the importance of modeling global dependencies is self-evident since there are distinct performance degradations on all the IQA metrics without the \emph{global branch}, e.g., 0.2446 dB in PSNR and 0.0026 in Q8 for the reduced-resolution test, and 0.0039 in $D_S$ for the full-resolution test, respectively.

\begin{figure}[t]
\centering
\includegraphics[width=0.805\linewidth]{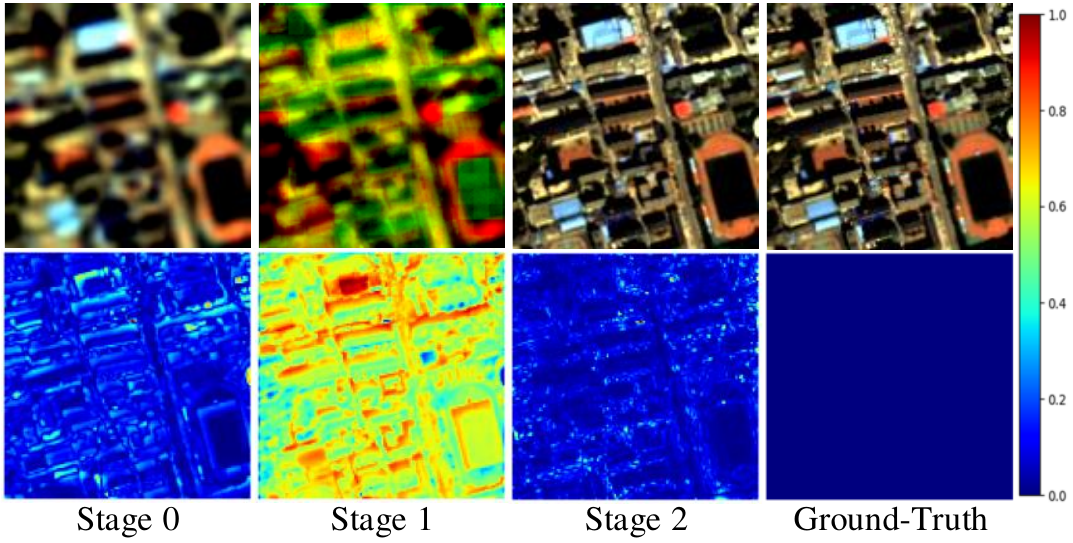}
% \includesvg[width=0.805\linewidth]{img/exper_visual/stage_visual.svg}
\vspace{-0.2cm}
\centering
\caption{Stage-wise visualization on the GaoFen-2 satellite scene.}
\label{fig4}
\end{figure}

\paragraph{Stage-wise Visualization.} As illustrated in Fig. \ref{fig4}, for our LGTEUN, we visualize the intermediate results of different stages ($\bar{\mathbf{Z}}_0$, $\bar{\mathbf{Z}}_1$, and $\bar{\mathbf{Z}}_2$) from a representative sample in the GaoFen-2 satellite data set, including the paired pan-sharpening and residual images. It is clear that more detailed information is recovered with LGTEUN iterating.

\paragraph{Limitations.} In short, two-fold potential limitations of our LGTEUN are as follows: \textbf{1)} The pan-sharpening results on the full-resolution scene have room for performance boosting. \textbf{2)} Further enhancements on model efficiency would make our proposed LGTEUN more competitive.

\section{Conclusion}
In this paper, for the MS pan-sharpening, we develop our LGTEUN by unfolding the designed PGD optimization algorithm into a deep network to improve the model interpretability. In our LGTEUN, to complement the lightweight data module, we customize a LGT module as a powerful prior module for image denoising to simultaneously capture local and global dependencies. To the best of our knowledge, LGTEUN is the first transformer-based DUN for the MS pan-sharpening, and LGT is also the first transformer module to perform spatial and frequency dual-domain learning. Comprehensive experimental results on three satellite data sets demonstrate the effectiveness and efficiency of our LGTEUN compared with other SOTA methods. 

\section*{Acknowledgments}
This work was supported in part by the National Natural Science Foundation of China under Grant U1903127 and in part by the Taishan Industrial Experts Programme under Grant tscy20200303.

\bibliographystyle{named}
\bibliography{res}

\end{document}